\documentclass[runningheads]{llncs}
\usepackage{graphicx}
\usepackage{amsmath,amssymb} 
\usepackage{color}
\usepackage{subfigure}
\usepackage{graphicx}
\usepackage{float}
\usepackage{wrapfig}
\usepackage{ulem}
\usepackage{varwidth}
\usepackage{capt-of}
\usepackage{multirow}
\begin{document}
\pagestyle{headings}
\mainmatter

\title{Non-local NetVLAD Encoding for \\Video Classification} 



\author{Yongyi Tang$^\dagger$\quad Xing Zhang$^\ddagger$\quad Jingwen Wang$^\dagger$\quad Shaoxiang Chen$^\ddagger$\quad \\Lin Ma$^\dagger$\qquad Yu-Gang Jiang$^\ddagger$}
\institute{$^\dagger$Tencent AI Lab\qquad$^\ddagger$Fudan University\\
\email{\{yongyi.tang92,skyezx2018,jaywongjaywong,forwchen\}@gmail.com\\
forest.linma@gmail.com\qquad ygj@fudan.edu.cn}}

\maketitle

\begin{abstract}
This paper describes our solution for the 2$^\text{nd}$ YouTube-8M video understanding challenge organized by Google AI. Unlike the video recognition benchmarks, such as Kinetics and Moments, the YouTube-8M challenge provides pre-extracted visual and audio features instead of raw videos. In this challenge, the submitted model is restricted to  1GB, which encourages participants focus on constructing one powerful single model rather than incorporating of the results from a bunch of models. Our system fuses six different sub-models into one single computational graph, which are categorized into three families. More specifically, the most effective family is the model with non-local operations following the NetVLAD encoding. The other two family models are Soft-BoF and GRU, respectively. In order to further boost single models performance, the model parameters of different checkpoints are averaged.
Experimental results demonstrate that our proposed system can effectively perform the video classification task, achieving 0.88763 on the  public test set and 0.88704 on the private set in terms of GAP@20, respectively. We finally ranked at the fourth place in the YouTube-8M video understanding challenge.
\end{abstract}

\section{Introduction}

Understanding video content is a major challenge for numeric applications including {video classification}~\cite{DBLP:conf/nips/SimonyanZ14}, video captioning~\cite{wang2018reconstruction,wang2018bidirectional}, video localization~\cite{chen2018temporally,feng2018video}, video attractiveness analysis~\cite{chen2018finegrained}, and so on. Especially with the exponential increment of online videos, video tagging, video retrieval and recommendation is of great demand. 
Therefore, developing reliable video understanding algorithms and systems has received extensive attentions in the area of computer vision and machine learning.

In order to recognize video content, convolutional neural networks(CNNs)~\cite{tran2015learning,DBLP:conf/nips/SimonyanZ14,carreira2017quo,qiu2017learning} and/or recurrent neural networks based methods~\cite{donahue2015long,tang2017latent} have achieved state-of-the-arts results. Those methods~\cite{DBLP:conf/nips/SimonyanZ14} take the advantages of deep learning methods on static image content as well as the video motion containing temporal information to perform video analysis. However, prior works only perform on those video benchmarks with limited number of videos for model evaluations such as UCF-101~\cite{soomro2012ucf101}, HMDB-51~\cite{jhuang2011large}, and ActivityNet~\cite{caba2015activitynet} datasets. Recently, several large-scale video datasets are constructed,  including the Kinetics dataset \cite{kay2017kinetics} developed by DeepMind and the Moments in Time dataset developed by MIT-IBM \cite{monfort2018moments} with about a million of videos. However, for practical video applications such as YouTube and Netflix, such number of videos is still relatively small and not suitable for large-scale video understanding. Nowadays, Google AI releases a large-scale video dataset named YouTube-8M~\cite{abu2016youtube}, which contains about 8 million YouTube videos with multiple class tags.

For the 1$^\text{st}$ Youtube-8M video understanding challenge, several techniques including context gating~\cite{miech2017learnable}, multi-stage training~\cite{wang2017monkeytyping}, temporal modeling~\cite{li2017temporal}, and feature aggregation~\cite{chen2017aggregating} have been proposed for video classification. However, the excellent performances of prior works mainly attribute to ensemble the results from a bunch of models, which is not practical in real-world applications due to the heavy computational expense. Therefore, the 2$^\text{nd}$ YouTube-8M video understanding challenge focus on learning video representation under budget constraints. More specifically, the model size of submission is restricted to 1GB, which encourages the participants to explore compact video understanding models based on
the pre-extracted visual and audio features.

In this report, we propose a compact system that meets the requirements and achieves superior results in the challenge. We summarize the contributions as follows.
First, we stack the non-local block with the NetVLAD to improve the video feature encoding. Experimental results demonstrate that the proposed non-local NetVLAD pooling method outperforms the vanilla NetVLAD pooling.
Second, several techniques are employed for building the large-scale video classification system with limited number of parameters including weight averaging strategy of different checkpoints, model ensemble, and compact encoding of floating point number.
Lastly, we show that the selected single models are complementary to each other which makes the whole system achieves a competitive result on the 2$^\text{nd}$ YouTube-8M video understanding challenge, ranked at the forth position.

\section{Approach}

\begin{figure*}[h]
\begin{center}
   \includegraphics[width=1.0\linewidth]{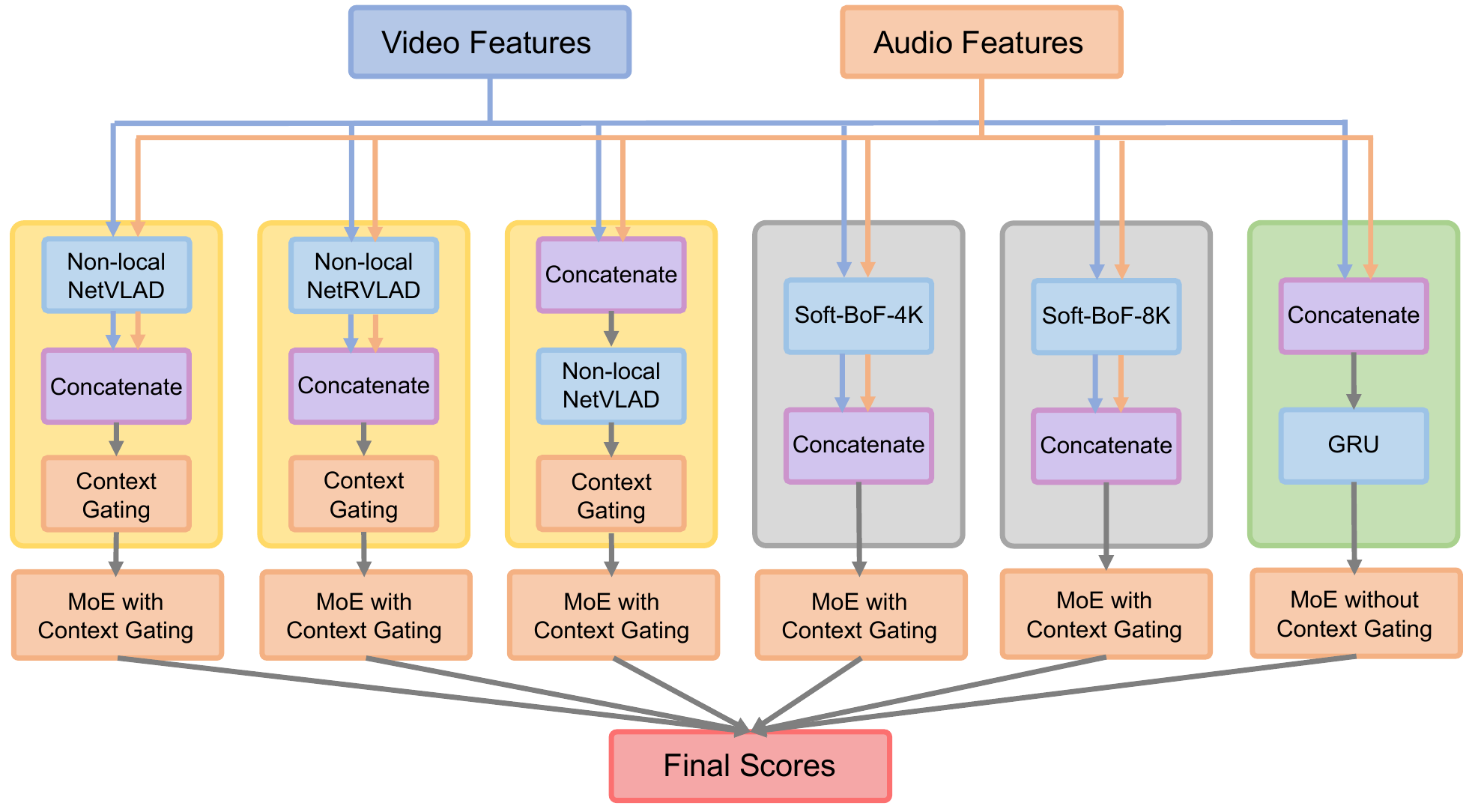}
\end{center}
   \captionof{figure}{The framework of our proposed system for video classification.}
    \label{fig:framework}
\end{figure*}

The framework of our proposed system is shown in Fig.~\ref{fig:framework}. In this work, we use three different families of video descriptor pooling methods for the video classification task, specifically the non-local NetVLAD, Soft-Bag-of-Feature (Soft-BoF), and GRU.
In section 2.1, we introduce the details of the proposed NetVLAD incorporated with the non-local block with its variants introduced in section 2.2. The other two family models, namely the Soft-BoF and GRU, are introduced in section 2.3 and 2.4, respectively. The model ensemble is described in section 2.5.

\subsection{Non-local NetVLAD}

\subsubsection{Vector of Locally Aggregated Descriptors (VLAD).} VLAD~\cite{jegou2010aggregating} is a popular descriptor pooling method for instance level retrieval~\cite{jegou2010aggregating} and image classification~\cite{gong2014multi}, as it captures the statistic information about the local descriptors aggregated over the image. Specifically, the VLAD summarizes the residuals of descriptors and its corresponding cluster center. Formally, given $N$ $D$-dimensional descriptors $\{\mathbf{x}_i\}$ as input, and $K$ cluster centers $\{\mathbf{c}_k\}$ as VLAD parameters, the pooling output of VLAD is $K\times D$-dimensional representation $V$. Writing $V$ as a $K\times D$ matrix, the $(j,k)$ element of $V$ can be computed as follows:
\begin{equation}
V(j,k) = \sum_{i=1}^N a_k(\mathbf{x}_i)(x_i(j)-c_k(j)),
\end{equation}
where the $a_k(\mathbf{x}_i)$ indicates the hard assignment of the descriptor $\mathbf{x}_i$ to $k$-th visual word $\mathbf{c}_k$. Thus, each column of matrix $V$ records the sum of residuals of the descriptors. Intra-normalization and inter-normalization are performed after VLAD pooling.

\subsubsection{NetVLAD Descriptor.}
However, the VLAD algorithm involves a hard cluster assignment that is non-differentiable. Thus the vanilla VLAD encoding is not appropriate for deep neural network that requires computing gradients for back-propagation. To address this problem, Arandjelovic \textit{et al.} proposed the NetVLAD \cite{arandjelovic2016netvlad} with soft assignment $\bar{a}_k(\mathbf{x}_i)$ of descriptors $\mathbf{x}_i$ to multiple clusters centers $\mathbf{c}_k$, \textit{i.e.},
\begin{equation}\label{eq:a_bar}
\bar{a}_k(\mathbf{x}_i) = \frac{e^{\mathbf{w}_k^T \mathbf{x}_i+b_k}}{\sum_{k'} e^{\mathbf{w}_{k'}^T \mathbf{x}_i+b_{k'}}},
\end{equation}
where $\{\mathbf{w}_k\}$, $\{b\}$ and $\{\mathbf{c}_k\}$ are the learnable  parameters of the NetVLAD descriptor.

\subsubsection{Non-local NetVLAD Descriptor.}
As described above, the VLAD descriptor uses cluster centers $\mathbf{c}_k$ to represent features, while NetVLAD further uses soft-assignment to construct the local feature descriptors. To enrich the information of NetVLAD descriptors, we model the relations between different local cluster centers. We employ the non-local block proposed by Wang \textit{et al.}~\cite{Wang_2018_CVPR}, which has already demonstrated the relation modeling ability in action recognition task. Here, we empirically adopt the embedded Gaussian function to compute the non-local relations:
\begin{equation}
f(\mathbf{v}_i,\mathbf{v}_j) = e^{\theta(\mathbf{v}_i)^T \phi(\mathbf{v}_j)}.
\end{equation}
Specifically, given the NetVLAD descriptor $\mathbf{v}_k$ corresponding to cluster centers $\mathbf{c}_k$, the non-local NetVLAD descriptor $\hat{\mathbf{v}}_k$ of cluster $k$ is formulated as:
\begin{equation}
\hat{\mathbf{v}}_i = \mathbf{W}\mathbf{y}_i + \mathbf{v}_i,
\end{equation}
where $\mathbf{y}_i = \frac{1}{Z(\mathbf{v})}\sum_{\forall j}f(\mathbf{v}_i,\mathbf{v}_j)g(\mathbf{v}_j)$. For implementation, the non-local NetVLAD is formulated as:
\begin{equation}
\mathbf{y}=\text{softmax}(\mathbf{v}^T\mathbf{W}_\theta^T\mathbf{W}_\phi\mathbf{v})g(\mathbf{v}),
\end{equation}
where $g(\mathbf{v})$ is a linear transformation.

\subsection{Non-local NetVLAD Model and its Variants}
Note that in our system, we use three variant non-local NetVLAD methods, which are demonstrated to be complementary with each other.

\subsubsection{Late-fused Non-local NetVLAD (LFNL-NetVLAD).}
The first model is the late-fused non-local NetVLAD (LFNL-NetVLAD). The pre-extracted visual feature and audio feature are encode independently by the non-local NetVLAD pooling method. Afterwards, these two non-local NetVLAD features, encoding visual and audio  modalities, are concatenated into a vector, which is followed by the context gating module.

Please note that context gating is introduced by Miech \textit{et al.}~\cite{miech2017learnable}, which transforms the input feature into a new representation and captures feature dependencies and prior structure of output space. Context gating is defined as:
\begin{equation}
\mathbf{z}=sigmoid(\mathbf{W}\mathbf{y})\odot\mathbf{y},
\end{equation}
where $\odot$ indicates elements-wise multiplication.

As shown in Fig.\ref{fig:framework}, the mixture of experts (MoE) model~\cite{jordan1994hierarchical} equipped with video level context gating is used for the multi-label video classification. 

\subsubsection{Late-fused Non-local NetRVLAD (LFNL-NetRVLAD).}
In addition, the NetRVLAD that drops the computation of cluster centers is proposed in~\cite{miech2017learnable}, which can be considered as self-attended local feature representation. 
Formally, the NetRVLAD can be defined as:
\begin{equation}
V'(j,k) = \sum_{i=1}^N \bar{a}_k(\mathbf{x}_i)x_i(j),
\end{equation}
where the soft assignment $\{\bar{a}_k\}$ are computed by Eq.~\ref{eq:a_bar}.
Similarly, the video and audio features pass through non-local NetRVLAD pooling and perform concatenation, followed by one context gating module and the MoE equipped with video level context gating. 

\subsubsection{Early-fused Non-local NetVLAD (EFNL-NetVLAD).}
Early fusion that concatenates the video and audio feature before non-local NetVLAD pooling is used to build another model. The early-fused feature lies in different feature space resulting in different expressive ability compared with the late-fused representation. The frame level context gating and video level MoE with context gating are also used in this model.

\subsection{Soft-Bag-of-Feature Pooling}
For bag-of-feature encoding, we utilize soft-assignment of descriptors to feature clusters~\cite{philbin2008lost} to obtain the distinguishable representation. Also, we perform late fusion of Soft-BoF with 4K and 8K clusters, which are named as Soft-BoF-4K and Soft-BoF-8K, respectively. Those outputs only followed by the video level MoE with context gating.

\subsection{Gated Recurrent Unit}
Recurrent neural networks, especially the Gated Recurrent Unit (GRU)~\cite{cho2014properties}, have been investigated for video understanding~\cite{donahue2015long,miech2017learnable,li2017temporal,chen2017aggregating}. We stacked two layers of GRU of 1024 hidden neurons for each layer. The experimental results demonstrate that the GRU model is complementary with the non-local NetVLAD and Soft-BoF families resulting a significant improvement after model ensemble.

\subsection{Model Ensemble}
Model ensemble is a common way for boosting final results in different challenges~\cite{miech2017learnable,chen2017aggregating,wang2017monkeytyping,li2017temporal,zhang2018}. The superior improvement may attribute to the various feature expressions of different models.
Thus, model ensemble helps to finalize a robust result and relief over-fitting. We perform model ensemble based on the six different models as mentioned. Experimental results along with implementation details will be introduced in the following.

\section{Experiments}

\subsection{YouTube-8M Dataset}
The YouTube-8M dataset~\cite{abu2016youtube} adopted in the 2nd YouTube-8M Video Understanding Challenge is the 2018 version with higher-quality, more topical annotations, and a cleaner annotation vocabulary. It contains about 6.1 million videos, 3862 class labels and  3 labels per video on average. Because of the large scale of the dataset, the video information is provided as pre-extracted visual and audio features at 1 FPS. 

\begin{table}[t]
  \captionof{table}{Single model performances on our split validation set.}
    \centering
    \resizebox{1.0\textwidth}{!}{
    {
    \begin{tabular}{|c|c|c|c|}
    \hline
    Model & LFNL-NetVLAD & EFNL-NetVLAD & LFNL-NetRVLAD \\
    \hline
    GAP@20 & 0.8703 & 0.8674 & 0.8687 \\
    \hline
    Model size & 593M & 427M & 478M \\
    \hline
    \hline
    Model & Soft-BoF-4K & Soft-BoF-8K & GRU \\
    \hline
    GAP@20 & 0.8525 & 0.8512 & 0.8568 \\
    \hline
    Model size & 109M & 143M & 243M\\
    \hline
    \end{tabular}
    }}
    
    \label{Tab:single_model}
\end{table}

\begin{table}[t]
  \captionof{table}{Single averaged model performances on our split validation set.}
    \centering
    \resizebox{1.0\textwidth}{!}{
    {
    \begin{tabular}{|c|c|c|c|}
    \hline
    Averaged Model & LFNL-NetVLAD & LFNL-NetRVLAD & EFNL-NetVLAD\\
    \hline
    GAP@20 & 0.8716 & 0.8704 & 0.8704 \\
    \hline
    Averaged Model & Soft-BoF-4K & Soft-BoF-8K & GRU\\
    \hline
    GAP@20 & 0.8574 & 0.8563 & 0.8612  \\
    \hline
    \end{tabular}
    }}
    
    \label{Tab:avg_model}
\end{table}

\subsection{Implementation Details}

The provided dataset is divided into training, validation and test subsets with around 70\%, 12\% and 18\% of videos. But in our work, we keep around 100K videos for validation, and the remaining videos of training and validation subset are used for training due to the observation of improvement. We found that the performance on our validation set was 0.02-0.03 lower than the test set on the public leader board. 
We report the Global Average Precision (GAP) metric at top 20 with our split validation subset and the public test set shown on the leader board.

For most of the models, we empirically used 1024 hidden states except for the GRU model which adopted 1200 hidden states.
We trained every single model independently with our training split on Tensorflow~\cite{abadi2016tensorflow}. The Adam optimizer\cite{kingma2014adam} with 0.0002 as the initial learning rate was employed throughout our experiments. 
Training procedures converged around 300k.
After finishing the training procedure, we built a large computational graph of model ensemble, and the parameters within this graph were imported from the independent models. The averaged score of each sub-model was the final score of our system. Further fine-tuning for the system may improve the final score. In the submission, we simply used model-wise averaging due to the lack of time.

\subsection{Single Model Evaluation}
In this section, we evaluate the six single models used in our system as shown in Table~\ref{Tab:single_model}. 
For the LFNL-NetVLAD model, we deployed 64 clusters with 8 MoE in video level model achieving 0.8702 GAP@20 in our validation set, while the vanilla NetVLAD achieves 0.8698 under the same settings.
Also, 64 clusters were adopted in the EFNL-NetVLAD and the LFNL-NetRVLAD since we found this setting keeps the balance between model size and performance. And the MoE of these two models were 2 and 4, respectively. The model size of non-local NetVLAD models are around 500M, which takes a large portion of the parameters in our system.

We also adopted the GRU model with model size as 243M and two smaller Soft-BoF models with 4K and 8K clusters, respectively, since we found that those models are complementary to the non-local NetVLAD models. The MoE of these three models were set to 2.

In order to further boost the single model performance, we employed linear model averaging that utilizing the average of multiple checkpoint to improve single model performance inspired by Stochastic Weight Averaging method~\cite{izmailov2018averaging}. The final GAP@20 of each model is shown in Table~\ref{Tab:avg_model}, which shows that linear model averaging can significantly improve single model performance especially for the GRU and Soft-BoF models with over 0.005 improvements.

\subsection{Tricks for Compact Model Ensemble}
Recall that the challenge requires less than 1GB model for final submission. We thus adopted several techniques for improving model abilities under the limited parameters including using 'bfloat16' format of parameters and repeatedly random sampling.

At first, we trained the network with float32 in Tensorflow~\cite{abadi2016tensorflow}, which means that it takes 4 bytes for every parameter. To make our model meet the model size requirement, we used a tensorflow-specific format, 'bfloat16', in the ensemble stage, which is different from IEEE's float16 format.
The bfloat16 is a compact 16-bit encoding of floating point number with 8 bits for exponent and 7 bits for mantissa. We found that using 'bfloat16' format can accelerate the process without significant performance decrease, with its benefits on halving the model size which makes ensembling multiple models become possible. As results, we performed ensemble with the models mentioned in Table~\ref{Tab:avg_model} into one computational graph as our final model as shown in Table~\ref{Tab:ensemble_model}. 

\begin{table}[t]
  \captionof{table}{Ensemble model performances on our split validation set. M1-M6 denote LFNL-NetVLAD, LFNL-NetRVLAD, EFNL-NetVLAD, Soft-BoF-4k, Soft-BoF-8k and GRU, respectively.}
    \centering
    \resizebox{1.0\textwidth}{!}{
    {
    \begin{tabular}{|c|c|c|}
    \hline
    Ensemble Model & Validation GAP@20 & Public-Test GAP@20 \\
    \hline
    M1 \& M4 & 0.8752 & - \\
    \hline
    M1 \& M2 & 0.8778 & - \\
    \hline
    M1 \& M6 & 0.8782 & 0.8790 \\
    \hline
    M1 \& M4 \& M6 & 0.8800 & - \\
    \hline
    M1 \& M2 \& M4 \& M6 & 0.8820 & - \\
    \hline
    M1 \& M2 \& M3 \& M4 \& M6 & 0.8839 & 0.88678 \\
    \hline
    M1 \& M2 \& M3 \& M4 \& M5 \& M6 & 0.8842 & - \\
    \hline
    \end{tabular}
    }}
    
    \label{Tab:ensemble_model}
\end{table}

Further, since feature sub-sampling were used in our sub-models for better generalization, we performed multiple running with different feature sub-sampling in the same system to produce the final classification result. By averaging the 10 times repeated results, the final performance gained about 0.0005 inprovement on our validation set as shown in Table~\ref{Tab:random_model}. In practice, we repeated the input feature several times, and averaged the results for each video.
The final model size of our submission is 995M.

\begin{table}[t]
  \captionof{table}{Performances of our model with different times of random averaging.}
    \centering
    \resizebox{1\textwidth}{!}{
    {
    \begin{tabular}{|c|c|c|}
    \hline
    Ensemble Model & Validation GAP@20 & Public-Test GAP@20 \\
    \hline
    Our model run once & 0.8842 & - \\
    \hline
    Our model run 5 times & 0.8846 & 0.88756 \\
    \hline
    Our model run 10 times (final submission) & 0.8847 & 0.88763 \\
    \hline
    \end{tabular}
    }}
    
    \label{Tab:random_model}
\end{table}

\section{Conclusions}

In this report, we proposed a compact large-scale video understanding system that effectively performs multi-label classification on the YouTube-8M video dataset with limited model size under 1GB. A non-local NetVLAD pooling method is proposed for constructing more representative video descriptors. Several models including LFNL-NetVLAD, LFNL-NetRVLAD, EFNL-NetVLAD, GRU, Soft-BoF-4K, and Soft-BoF-8K are incorporated in our system for model ensemble. To halve model size, bfloat16 format is adopted in our final system. Averaging multiple outputs after random sampling is also used in our system for further boosting the performance. Experimental results on the 2nd YouTube-8M video understanding challenge show that the proposed system outperforms most of the competitors, ranking the fourth place in the final result.

\end{document}